# Heart Disease Prediction System using Associative Classification and Genetic Algorithm

M.Akhil jabbar [a]*, Dr.Priti Chandra[b], Dr.B.L Deekshatulu[c]

[a]*Research Scholar,JNTU Hyderabad,A.P INDIA*
[b]*Senior Scientist,Advanced System Laboratory,DRDO,,Hyderabad,INDIA*
[c] Distinguished fellow, IDRBT ,RBI,Govt of INDIA

**Abstract**

Associative classification is a recent and rewarding technique which integrates association rule mining and classification to a model for prediction and achieves maximum accuracy. Associative classifiers are especially fit to applications where maximum accuracy is desired to a model for prediction. There are many domains such as medical where the maximum accuracy of the model is desired. Heart disease is a single largest cause of death in developed countries and one of the main contributors to disease burden in developing countries. Mortality data from the registrar general of India shows that heart disease are a major cause of death in India, and in Andhra Pradesh coronary heart disease cause about 30%of deaths  in rural areas. Hence there is a need to develop a decision support system for predicting heart disease of a patient. In this paper we propose efficient associative classification algorithm using genetic approach for heart disease prediction. The main motivation for using genetic algorithm in the discovery of high level prediction rules is that the discovered rules are highly comprehensible, having high predictive accuracy and of high interestingness values. Experimental Results show that most of the classifier rules help in the best prediction of heart disease which even helps doctors in their diagnosis decisions.

## 1. Introduction

The major reason that the data mining has attracted great deal of attention in the information industry in the recent years is due to the wide availability of huge amounts of data and the need for turning such data into useful information and knowledge. The information gained can be used for applications ranging from business management, production control, and market analysis to emerging design and science exploration and health data analysis [1]. Association rule mining and classification are two main functionalities of data mining. Association rule mining is used to find associations or correlations among the item sets. It is a unsupervised learning where no class attribute is involved in finding the association rule. On the other hand, classification is a supervised learning where class attribute is involved in the construction of the classifier and is used to classify or predict the data unknown sample.

---

* Corresponding author. Tel.: +91 9912648686
*E-mail address*: jabbar.meerja@gmail.com.

Associative classification involves two stages.
1) Generate class based association rules from a training data set
2) Classify the test data set into predefined class labels.

There is growing evidence that merging classification and association rule mining together can produce more efficient and accurate classification system than traditional classification techniques.

Genetic algorithms are typically used for problems that cannot be solved efficiently with traditional techniques. Genetic algorithms seem to be useful for searching very general spaces and optimization problems.

Coronary heart disease is epidemic in India and one of the major causes of disease burden and deaths. Data from registrar general of India shows that heart diseases are major cause of death in India, studies to determine the precise cause of death in urban Chennai and rural areas of A.P have revealed that CVD cause about 40% of the deaths in urban and 30% in rural areas [2].

In this paper we propose genetic algorithm based associative classification for heart disease prediction. A brief introduction about associative classification, genetic algorithms and heart diseases are given in the following subsections, followed by related work in section 2.section 3 deals with our proposed method. Section 4 deals with results and discussions. We will conclude our final remarks in section 5.

## 1.1  Associative Classification

Classification is one of the most important tasks in data mining. Researchers are focusing on designing classification algorithm to build accurate and efficient classifiers for large data sets. Associate classification achieves high accuracy, its rules are interpretable and it provides confidence probability when classifying objects which can be used to solve classification problem uncertainity.Therefore, it becomes hot theme in recent years [3].

Associative classification is a special case of association rule mining in which class attribute is considered in the rule's consequent. For example in a rule A$\rightarrow$B, B must be a class attribute. A classifier is of the form

$A_1$, $A_2$, ---$A_n$$\rightarrow$B, where $A_i$ is an attribute and B is a class. Rule item that satisfy minsup are called frequent rule items, while the rest are called infrequent rule items.Assocative classification is to collect rules in training data set D, organize them in a certain order to form a classifier. When provided an unlabelled object, the classifier selects the rule in accordance with the order whose condition matches the objects and assigns class labels of the rule to it.

**Table 1**.Training data set

| Sl.no | A1 | A2 | A3 | CLASS |
|-------|------|------|------|-------|
| 1 | a 11 | a 21 | a 31 | C1 |
| 2 | a 12 | a 24 | a 32 | C2 |
| 3 | a 13 | a 23 | a 33 | C0 |
| 4 | a 11 | a 21 | a 31 | C1 |
| 5 | a 12 | a 22 | a 32 | C2 |

## 1.2  Genetic Algorithm

Genetic algorithms are computing methodologies constructed in analogy with the process of evolution [4].It closely resembles the natural process of regeneration, reproduction, inheritance evolution. Genetic algorithms are typically used for problems that cannot be solved efficiently with traditional techniques. Genetic algorithms seem to be useful for searching very general spaces and optimization problems. Each solution generated in Genetic algorithms is called a chromosome (individual).Each chromosome is made up of genes, which are the individual elements (alleles) that represents the problem. The collection of chromosomes is called a population. The internal representation of the chromosomes is known as its genotype. This can be either bit strings or gray codes or hexadecimal codes. The external manifestation of the genotype or the real world representation of the genotype is



known as the phenotype [5].Basically there are three genetic operators are used for generating new strings. The functions of genetic operators are as follows:

1) Selection: selection deals with the probabilistic survival of the fittest in that, more fit chromosomes are chosen to survive.

2) Crossover: This operation is performed by selecting a random gene along the length of the chromosomes and swapping all the genes after that point. Various types of crossover operators are a) single point b) two point c) uniform d) half uniform e) reduced surrogate crossover f) shuffle crossover g) segmented crossover [6].

3) Mutation: mutation alters the new solutions so as to add stochasticity in the search for better solution. The most common method way of implementing mutations is to select a bit at random and flip (change) its value. There are 2 types of mutations use in genetic network programming 1) mutating the judgment node 2) mutating the value of the judgment node. In associative classification attributes and their values are taken as judgment nodes and class values as processing nodes.

Fitness function: Ideally the discoved rules should have a) high predictive accuracy b) be comprehensible c) be interesting. The accomplishment of a genetic algorithm is directly linked to the accuracy of the fitness function.

## 1.3   Heart Disease

Coronary heart disease is a narrowing of the small blood vessels that supply blood and oxygen to the heart. This is also called as coronary artery disease. Coronary heart disease is usually caused by a condition called atherosclerosis, which occurs when fatty material and a substance called plaque builds up on the walls of arteries. This causes them to get narrow. As the coronary arteries narrow, blood flow to the heart can slow down or stop, causing chest pain,shoreteness of breath, heart attack, and other symptoms. Men in their 40's have higher risk of Coronary heart disease than women, but as women gets older, their risk increases so that it is almost equal to a man's risk. Major risk factors for Coronary heart disease are 1) Diabetes    2) High blood pressure 3) High LDL (bad) cholesterol 4) low LDL (good) cholesterol 5) Not getting enough physical activity6) Obesity7) Smoking.

India is undergoing rapid epidemiological transition as a consequence of economic and social change, and cardiovascular disease is becoming an increasingly important cause of death. India's disease pattern has undergone a major shift over the past decade. As per WHO report, at present out of 10 deaths in India, eight are caused by non communicable diseases, such as cardio vascular diseases, and diabetes in urban india.In rural India, 6 out of every 10 deaths is caused by NCD'S [7].Data from registrar general of India shows that heart attacks are major cause of deaths in india.in Andhra Pradesh 30% of rural population died due to CHD.

There is an urgent need for development and implementation of suitable primordial, primary, and secondary prevention approaches to control this epidemic. An urgent and sincere bureaucratic, political, and social will to initiate steps in this direction is required.

## 2      Related Work

Large no. Of work is carried out in finding efficient methods of medical diagnosis for various diseases. Our work is an attempt to predict the cardiac disease in Andhra Pradesh using data mining.

Carlos implemented efficient search for diagnosis of heart disease comparing association rules with decision trees [8].A novel technique to develop the multi-parametric feature with linear and non linear characteristics of HRV was proposed by Hean Gyu lee et al.[9].A model intelligent heart diseases prediction system built with the aid of data mining techniques like decision trees, naive bayes and neural network was proposed by sellappan palaniappan et al[10]. The problem of identifying constrained association rules for heart disease prediction was studied by Carlos



Ordonez [11].MA.jabbar, Priti Chandra, B.L.Deekshatulu proposed evolutionary algorithm for heart disease prediction. They used genetic algorithm to predict the heart disease for Andhra Pradesh population [1].Enhanced prediction of heart disease with feature subset selection using genetic algorithm was proposed by M.Ambarasi et al [12].Intelligent and effective heart attack prediction system using data mining and AINN was proposed by [13].They employed the multilayer perception neural network with back propagation as the training algorithm. Graph based approach for heart disease prediction was proposed by MA.jabbar, B.L.Deekshatulu, and Priti Chandra [14].Their method is based on maximum clique and weighted association rule mining. Associative classification for heart disease prediction was proposed by MA.jabbar, B.L.Deekshatulu, and Priti Chandra [15].They used Gini index based classification to predict the heart disease. Cluster based association rule mining for heart attack prediction was proposed by MA.jabbar, B.L.Deekshatulu, and Priti Chandra [16].Their method is based on digit sequence and clustering. The entire data base is divided into partitions of equal size and association rule will be mined from each partition.

In this paper we propose efficient association classification for heart disease prediction for Andhra Pradesh population. We used Gini index to produce a compact rule set and filter rules further by applying Z-Statistics and genetic algorithm.

# 3    Proposed Method

Most of the associative classification algorithms adopt the exhaustive search method presented in the famous APRIORI algorithm to discover the rules and require multiple passes over the data base. Furthermore, they find frequent items in one phase and generate the rules in a separate phase consuming more resources such as storage and processing time. Moreover, since rule ranking plays an important role in classification and the majority of the associative classifiers select rules mainly in terms of their confidence levels. Even after pruning infrequent items, the APRIORI association rule generation procedure, produces a huge no. of association rules .If all the rules are used in the classifier then the accuracy of the classifier would be high but the building of classification will be slow. In order to improve the accuracy of associative classification we propose an informative attribute entered rule generation and hypothesis testing Z- statistics for heart disease prediction. The class association rules are represented as chromosomes and Michigan approach is used to encode the rules.

## 3.1   Proposed Algorithm

**STEP 1** :  find Gini index of each attribute. The attribute with minimum Gini index is selected for class association rule generation. These class association rules are known as initial population and represented as chromosomes.

$$Gini(t) = 1 - \sum_{i=0}^{c-1} [p(i/t)]^2$$

**STEP 2**: Evaluate fitness of rule using Z statistics

$$Z=S(X)-Minimum\ support/SQRT\ (min\ sup*(1-minsup))/N$$

Where **S(X)** is support of pattern and **min sup** is user defined threshold

**STEP 3**: Prune the rules based on Z statistics. After rule evaluation the rules having highest fitness are stored in a pool. Then apply genetic functions on these rules.

**STEP 4**: Perform single point cross over. Judgement nodes are selected for crossover.



**STEP 5**: Perform mutation by mutating the value of judgment node. This process will be repeated till last generation reached.

**STEP 6**: Build classifiers using the generated Rules

**STEP 7**: Predict the rules on test data

**STEP 8**: Find the accuracy of the data set

**Accuracy** = Number of objects correctly Classified
Total No. of objects in the test set.

## 3.2 Explanation of Algorithm

### A) Attribute selection based on Gini Index

An informative attribute centred rule generation produces a compact rule. **Gini** index is used as filter to reduce the no. of candidate item sets. It is used to select the best attribute. Those attributes with minimum Gini index are selected for rule generation.

$$Gini(t) = 1 - \sum_{i=0}^{c-1} [p(i/t)]^2$$

(1)

Let us consider a sample medical training data set given in table 2.

**Table2**: Example Medical Training data

| No. | mcv | alkphos | sgpt | sgot | gammagt | drinks | selector |
|-----|------|---------|------|------|---------|--------|----------|
| 1 | 85.0 | 92.0 | 45.0 | 27.0 | 31.0 | 0.0 | 1 |
| 2 | 85.0 | 64.0 | 59.0 | 32.0 | 23.0 | 0.0 | 2 |
| 3 | 86.0 | 54.0 | 33.0 | 16.0 | 54.0 | 0.0 | 2 |
| 4 | 91.0 | 78.0 | 34.0 | 24.0 | 36.0 | 0.0 | 2 |
| 5 | 87.0 | 70.0 | 12.0 | 28.0 | 10.0 | 0.0 | 2 |
| 6 | 98.0 | 55.0 | 13.0 | 17.0 | 17.0 | 0.0 | 2 |
| 7 | 88.0 | 62.0 | 20.0 | 17.0 | 9.0 | 0.5 | 1 |
| 8 | 88.0 | 67.0 | 21.0 | 11.0 | 11.0 | 0.5 | 1 |
| 9 | 92.0 | 54.0 | 22.0 | 20.0 | 7.0 | 0.5 | 1 |
| 10 | 90.0 | 60.0 | 25.0 | 19.0 | 5.0 | 0.5 | 1 |

After calculating Gini index of each attribute sgpt has the lowest Gini index. So sgpt would be the better attribute.

The rules generated like the following are considered for classifier.

1. sgpt='(-inf-19.1] ==> selector=2
2. sgpt='(-inf-19.1]' gammagt='(-inf-34.2]' ==> selector=2
3. sgpt='(19.1-34.2]' sgot='(20.4-28.1]' ==> selector=2
4. sgpt='(19.1-34.2]' gammagt='(-inf-34.2]' ==> selector=1
5. sgpt='(19.1-34.2]' drinks='(-inf-2]' ==> selector=1

### B) Z-Statistic (Hypothesis Testing)

Hypothesis Testing is a statistical inference procedure to determine whether a conjecture or hypothesis should be accepted or rejects based on the evidence gathered from data [17]. In our proposed approach we use Z-Statistic to verify the quality of pattern or rule.Various steps involved in testing of hypothesis'-Statistic is preferred if the sample size is greater than 30.

1) **Null Hypothesis**: Define a null hypothesis $H_O$ taking into consideration the nature of the problem and data involved.
2) **Alternative Hypothesis**: set up alternative hypothesis $H_1$ so that we could decide whether we should use one tailed and two tailed test.
3) **Level of significance**: Select the appropriate level of significance( )
4) **Test statistics:** compute Z statistics
   Z=S(X)-Minimum support/SQRT (min sup*(1-minsup))/N
5) **Conclusion** :compare the computed value of Z statistics with the critical value of Z (given in Z-Statistic Table) at given level of significance( )
   If |Z|< Z accept the null hypothesis
   If |Z|>| Z reject null hypothesis [18].Fig 1 shows left tail and right tail test.

Table 3: Critical values of Z

| *Level of significance* | *1%* | *5%* | *10%* |
|---|---|---|---|
| Two tailed test | \|Z \|=2.58 | \|Z \|=1.96 | \|Z \|=1.645 |
| Right tailed test | Z =2.33 | Z =1.645 | Z =1.28 |
| Left tailed test | Z =-2.33 | Z =-1.645 | Z =-1.28 |

**Example**: Let N=10000 S(X) =11% Minimum Support =10% Z-Statistic under null hypothesis is Z=3.33.Suppose level of significance ( ) =0.001 sets up a rejection region with Z =3.09. Since Z> Z  the null hypothesis is rejected and the pattern is considered statistically interesting.

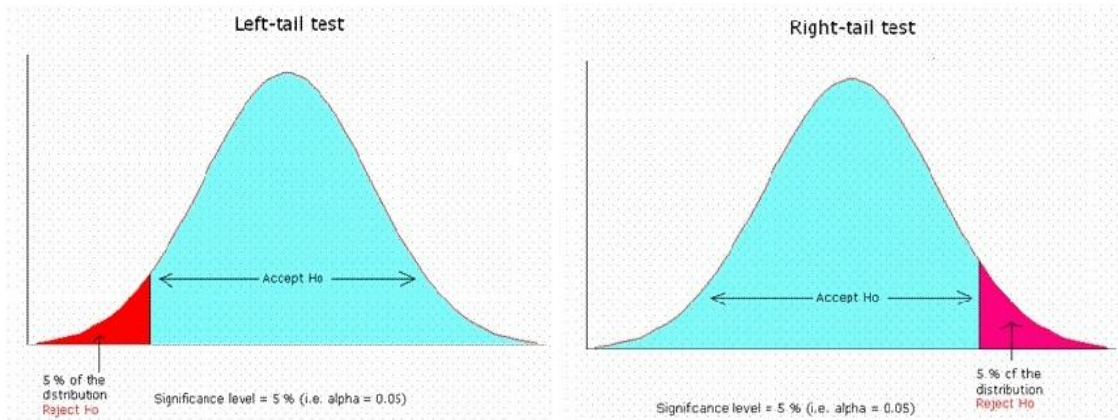

Fig. 1. (a) Left tail test                    (b) right tail test

### C) Crossover and Mutation

Crossover operator forms off springs by combining judgment nodes which are selected as crossover nodes. We used single point crossover in our approach

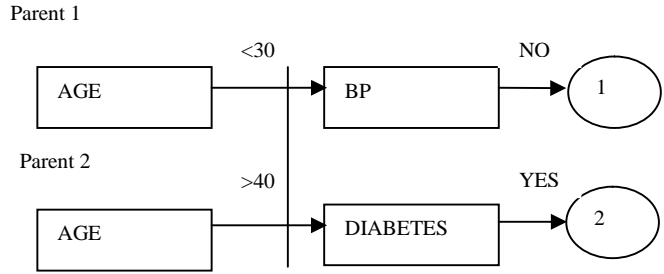

**After crossover**

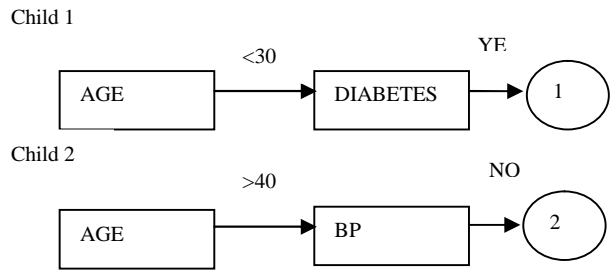

### 1) Mutation the judgment node

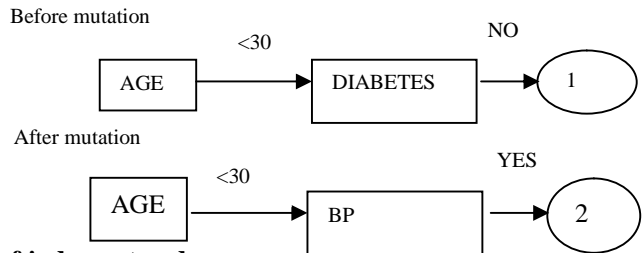

### 2) Mutation the value of judgment node

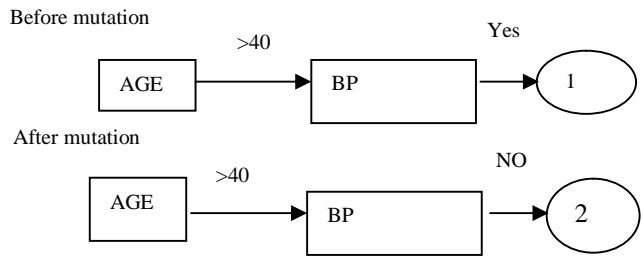

### D) Accuracy Computation

Accuracy measures the ability of the classifier to correctly classify unlabelled data.

**Accuracy** = Number of objects correctly Classified
Total No. of objects in the test set.

## 4    Results and Discussion

We have evaluated the accuracy of our proposed method on 6 data sets from SGI machine learning repository [19] and 2 medical data sets from UCI Machine learning repository [20].A brief description about the data sets was presented in table 4.Attributes selected based on Gini index for various data sets is shown in table 5.The accuracy is obtained using 10 fold cross validation. Table 6 shows the accuracy for the different data sets using our proposed approach. Figure 1&3 shows accuracy of various data sets. Table 8 and 9 represents comparison of accuracy on various data sets.fig 2 shows statistics of heart disease.

**Table 4**: Data set Description

| Data Sets | Transactions | Items | Classes |
|---|---|---|---|
| XD6 Data | 150 | 9 | 2 |
| Parity | 100 | 10 | 2 |
| Lens Data | 24 | 9 | 3 |
| Multiplexer Data | 100 | 12 | 2 |
| Weather Data | 14 | 5 | 2 |
| Balloon data | 36 | 4 | 2 |
| Diabetes | 768 | 9 | 2 |
| Breast cancer | 286 | 9 | 2 |

**Table 5**: Data sets and attribute selected based on Gini index

| Data set | Attribute selected based on Gini index |
|---|---|
| XD6 Data | A8 |
| Parity data | A5 |
| multiplexer | OUTPUT 1 |
| Lens data | Tear production rate |
| Balloon data | Age |
| Weather data | Humidity |
| Diabetes | Plas |
| Breast cancer | Deg lalig |

**Table 6**: Accuracy of various data sets

| Data set | Accuracy |
|---|---|
| XD6 Data | 75.75 |
| Parity data | 72 |
| multiplexer | 60 |
| Lens data | 84 |
| Balloon data | 84 |
| Weather data | 92.8 |

**Table 7:** Attributes selected for heart disease prediction

| Sl.no | Attribute Name |
|---|---|
| 1 | Age |
| 2 | BP Systolic |
| 3 | BP Diastolic |
| 4 | gender |
| 5 | Hypertension |
| 6 | Diabetes |
| 7 | Rural/Urban |

**Table 8**: Accuracy of medical data sets

| Data sets | J48 | Naïve bayes | GNP | NN | Our Method | GNP Using chi square method |
|---|---|---|---|---|---|---|
| Pima Indian Diabetes | 75.5 | 76.3 | 78 | 65.1 | 82 | 74.41 |
| Breast Cancer | 74.2 | 71.67 | 77.27 | 70.27 | 84.8 | 93 |
| **Heart Disease Data (A.P)** | 80 | 76 | - | 82 | 98 | - |
| **Average accuracy** | **76.56** | **74.65** | **77.6** | **72.45** | **88.9** | **83.70** |

**Table 9**: Accuracy of Data Sets (non medical)

| Data sets | C4.5 | Naïve bayes | Our Proposed Method |
|---|---|---|---|
| XD6 | 78.6 | - | 76 |
| Parity | 53.3 | 40 | 72 |
| Lens | 83.3 | 70.8 | 84 |
| Multiplexer | 61 | 61.9 | 60 |
| Balloon | 83 | 72 | 84 |
| Weather | 50 | 57.14 | 92.8 |

The performance of our proposed method is evaluated on 2 medical data sets diabetes and cancer data sets by comparing it with the traditional classification algorithms like j48, naive bayes, neural networks and GNP [21].Accuracy of pima data has been improved using genetic network programming (GNP).it has 1%improvement than traditional naive bayes classification algorithm, and 4.6%improvement over breast cancer data. The accuracy of heart disease data using j48 is 4% higher than naive bayes and the accuracy has been improved by using NN.J48 outperformed naive bayes and neural networks for Pima and cancer data. Our proposed approach reached 7.5%improvement over GNP with chi square[22]for pima Indian diabetes data. Our approach reached the best accuracy, compared with other classification algorithms. So our proposed algorithm performs better than traditional classification algorithms. A possible reason for efficient classification system produced by our algorithm is the fact that it employs genetic algorithm and hypothesis testing. The use of genetic algorithm and hypothesis testing Z Statistic proposed in our algorithm suggest the validity of the hypothesis, which states the more constraints imposed to discriminate between generated rules, the more random selection is minimized, which increases the accuracy of the classifier.

The data for heart disease prediction was collected from various corporate hospitals from Andhra Pradesh and opinion from expert doctors. Attribute age is discritized as age 0-45 and age>45.Attribute age>45 is chosen as information centred attribute based on computing Gini index. Some of the rules generated predicting the Heart diseases are

1. AGE>45, BP Diastolic, BP systolic, diabetes=> Heart Disease
2. Age>45, BP Diastolic, BP systolic, Hypertension, diabetes=> Heart Disease
3. AGE>45, BP Diastolic, diabetes=> Heart Disease
4. AGE>45, BP Diastolic, Male, diabetes=> Heart Disease
5. AGE>45, BP Diastolic, Male, Hypertension, diabetes=> Heart Disease
6. AGE>45, BP Diastolic, Hypertension, diabetes=> Heart Disease
7. AGE>45, BP Diastolic, Hypertension, rural=> Heart Disease
8. AGE>45, BP systolic, Hypertension, diabetes=> Heart Disease
9. AGE>45, diabetes, rural=> Heart Disease
10. AGE>45, Male, Hypertension, diabetes=> Heart Disease
11. AGE>45, Male, Hypertension, rural=> Heart Disease
12. AGE>45, Hypertension, diabetes=> Heart Disease

**Results:**
1) Majority of the people who had CVD were in the age group 46-65
2) Among all the participants of the study 60%of the males and 40%of the females had heart disease.
3) 50%of the males who had hypertension are associated with CVD
4) 8%of the females who had hypertension are associated with CVD
5) A Higher percentage of males were found to be Diabetic (30%)
6) 38% of the people who lives in urban areas are associated with heart disease
7) Hypertension and Diabetes account for 30%of all cases.
8) Among all the cases males had a higher systolic pressure(44% cases)
9) 32%of Males who live in urban areas are associated more with heart disease.

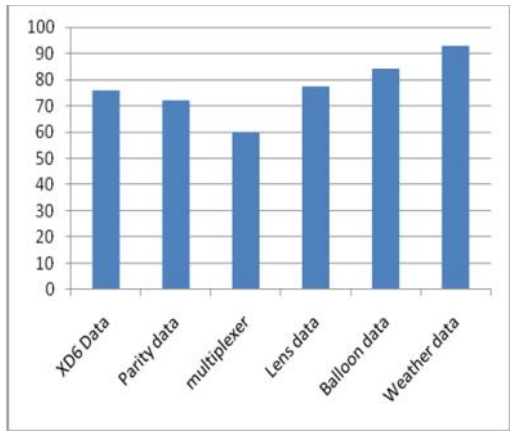
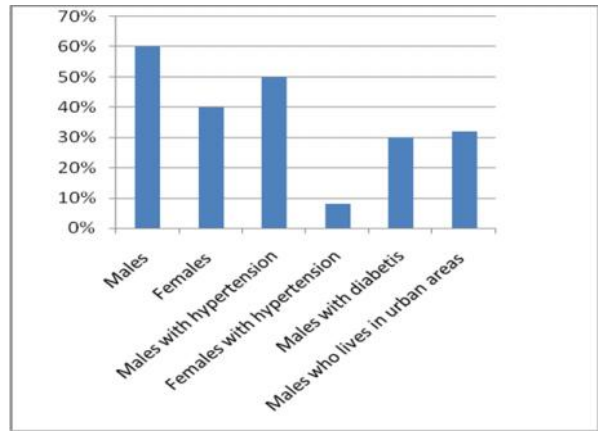

**Fig. 1** Accuracy of various data sets by our algorithm     **Fig. 2** Statistics of Heart Disease

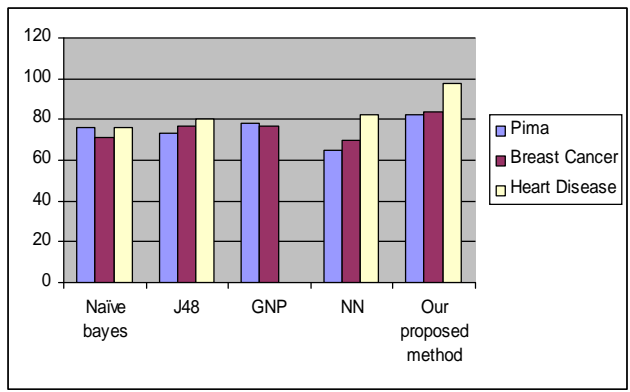

**Fig .3** Accuracy of Medical data sets

## 5    Conclusion And Future work

In the recent years India and other developing countries have witnessed a rapidly escalating epidemic of cardiovascular disease (CVD).It is predicted that by 2020 coronary heart disease will be leading cause of death in adult Indians, and Andhra Pradesh is in risk of more deaths due to CVD.The need to contain the epidemic of cardiovascular disease as well as combat its impact and minimize its toll on Andhra Pradesh is obvious and urgent. Hence a decision support system is proposed to identify a risk score for predicting the heart disease. In this paper, we proposed a system for heart disease prediction using data mining techniques. In our feature work we plan to reduce no. of attributes and to determine the attribute which contribute towards the diagnosis of disease using genetic algorithm.